\documentclass[10pt,twocolumn,letterpaper]{article}

\usepackage{cvpr}
\usepackage{times}
\usepackage{epsfig}
\usepackage{graphicx}
\usepackage{amsmath}
\usepackage{amssymb}
\usepackage{bbm}

\usepackage[pagebackref=true,breaklinks=true,letterpaper=true,colorlinks,bookmarks=false]{hyperref}

\cvprfinalcopy 

\def\cvprPaperID{1303} 

\ifcvprfinal\fi
\begin{document}

\title{Instance Segmentation by Jointly Optimizing Spatial Embeddings and Clustering Bandwidth}

\author{Davy Neven \quad Bert De Brabandere \quad Marc Proesmans \quad Luc Van Gool\\
	Dept.\ ESAT, Center for Processing Speech and Images\\
 	KU Leuven, Belgium\\
{\tt\small \{firstname.lastname\}@esat.kuleuven.be} \\
}

\maketitle

\begin{abstract}
Current state-of-the-art instance segmentation methods are not suited for real-time applications like autonomous driving, which require fast execution times at high accuracy. Although the currently dominant proposal-based methods have high accuracy, they are slow and generate masks at a fixed and low resolution. Proposal-free methods, by contrast, can generate masks at high resolution and are often faster, but fail to reach the same accuracy as the proposal-based methods. In this work we propose a new clustering loss function for proposal-free instance segmentation. The loss function pulls the spatial embeddings of pixels belonging to the same instance together and jointly learns an instance-specific clustering bandwidth, maximizing the intersection-over-union of the resulting instance mask.
When combined with a fast architecture, the network can perform instance segmentation in real-time while maintaining a high accuracy. We evaluate our method on the challenging Cityscapes benchmark and achieve top results (5\% improvement over Mask R-CNN) at more than 10 fps on 2MP images.
Code will be available at: \url{https://github.com/davyneven/SpatialEmbeddings}
\end{abstract}

\section{Introduction}
Semantic instance segmentation is the task of locating all objects in an image, assigning each object to a specific class and generating a pixel-perfect mask for each one, perfectly delineating their shape. This contrasts with the standard bounding-box detection methods, where each object is represented by a crude rectangular box. Since having a binary mask for each object is desired (and necessary) in many applications, ranging from autonomous driving and robotics applications to photo-editing/analyzing applications, instance segmentation remains an important research topic.

\begin{figure}[t]
    \begin{center}
    	\includegraphics[width=1\linewidth]{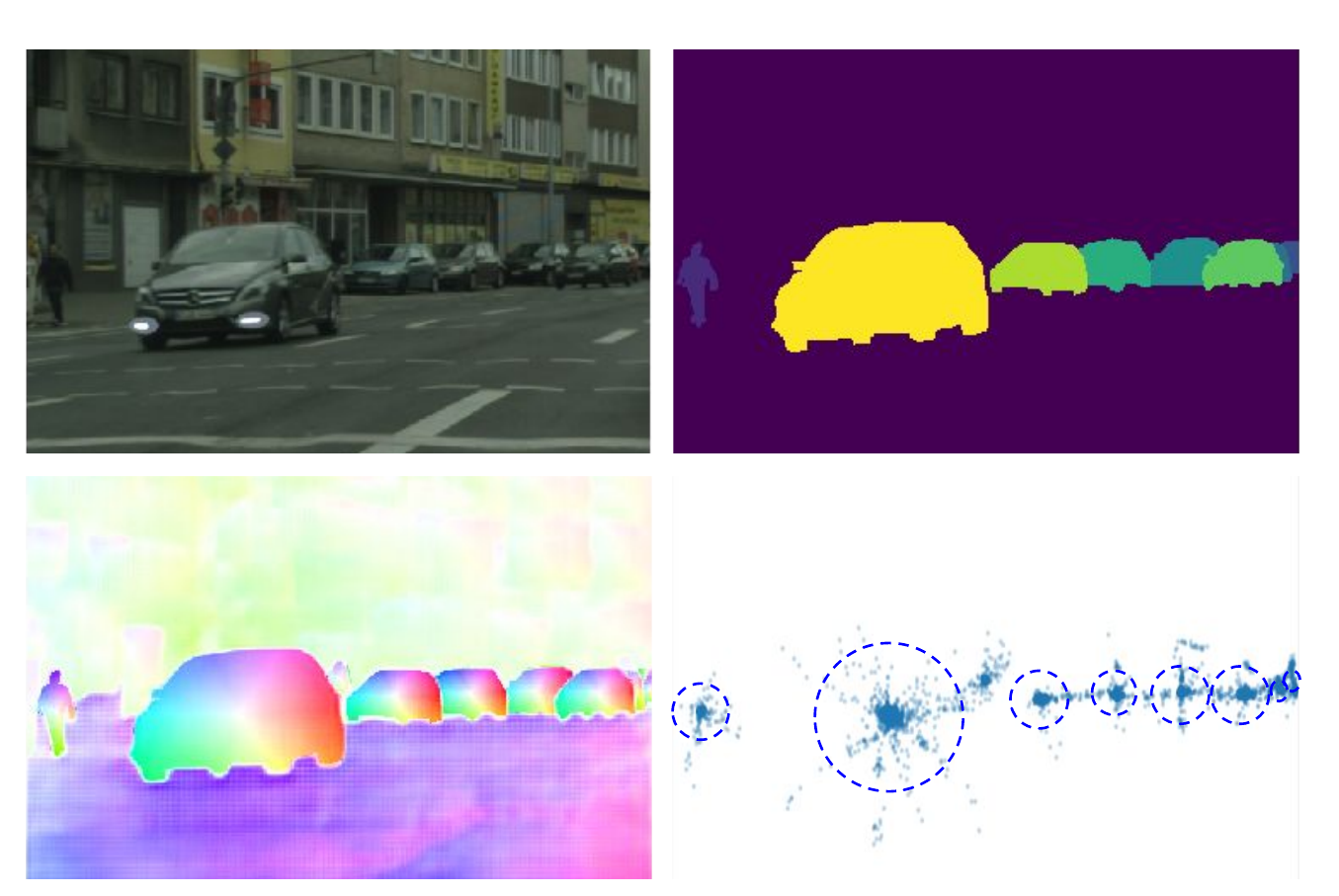}
    \end{center}
    \caption{Our loss function encourages pixels to point into an optimal, object-specific region around the object's center, maximizing the intersection-over-union of each object's mask. For big objects, this region will be bigger, relaxing the loss for edge-pixels, which are further away from the center. Bottom left displays the learned offset vectors, encoded in color. Bottom right displays the displaced pixels, displaced with the learned offset vectors. Instances are recovered by clustering around each center with the learned, optimal clustering region.}
    \label{fig:teaser}
\end{figure}

Currently, the dominant method for instance segmentation is based on a detect-and-segment approach, where objects are detected using a bounding-box detection method and then a binary mask is generated for each one. 
Despite many attempts in the past, the Mask R-CNN framework was the first one to achieve outstanding results on many benchmarks, and is still the most used method for instance segmentation to date. While this method provides good results in terms of accuracy, it generates low resolution masks which are not always desirable (e.g.~for photo-editing applications) and operates at a low frame rate, making it impractical for real-time applications such as autonomous driving.

Another popular branch of instance segmentation methods are proposal-free methods, which are mostly based on embedding loss functions or pixel affinity learning. Since these methods typically rely on dense-prediction networks, their generated instance masks can have a high resolution.
Additionally, proposal-free methods often report faster run-times than proposal-based ones.
Although these methods are promising, they fail to perform as well as the above mentioned detect-and-segment approaches like Mask R-CNN.

In this paper, we formulate a new loss function for proposal-free instance segmentation, combining the benefits of both worlds: accurate, high resolution masks combined with real-time performance. Our method is based on the principle that pixels can be associated with an object by pointing to that object's center. Unlike previous works that apply a standard regression loss on all pixels, forcing them to point directly at the object's center, we introduce a new loss function which optimizes the intersection-over-union of each object's mask. Our loss function will therefore indirectly force object pixels to point into an optimal region around the object's center. For big objects, the network will learn to make this region bigger, relaxing the loss on pixels which are further away from the object's center. At inference time, instances are recovered by clustering around each object's center with the learned, object-specific region. See figure~\ref{fig:teaser}.

We test our method on the challenging Cityscapes dataset and show that we achieve top results, surpassing Mask R-CNN with an Average Precision score of 27.6 versus 26.2, at a frame rate of more than 10 fps. We also observe that our method does very well on cars and pedestrians, reaching similar accuracy scores as a Mask R-CNN model which was trained on a combination of Cityscapes and COCO. On the Cityscapes dataset, our method is the first one which runs in real time while maintaining a high accuracy.

In summary, we (1) propose a new loss function which directly optimizes the intersection-over-union of each instance by pulling pixels into an optimal, object-specific clustering region and (2) achieve top results in real-time on the Cityscapes dataset.

\section{Related Work}
The current best performing instance segmentation methods are proposal-based, and rely on the Faster R-CNN~\cite{ren2015faster} object detection framework, which is the current leader in most object detection benchmarks. Previous instance segmentation approaches relied on their detection output to get object proposals, which they then refine into instance masks~\cite{dai2016instance,li2016fully,pinheiro2015learning,pinheiro2016learning}. Mask R-CNN~\cite{he2017mask} and its derivative PANet~\cite{liu2018path} refine and simplify this pipeline by augmenting the Faster R-CNN network with a branch for predicting an object mask. Although they are the best-scoring methods on popular benchmarks, such as COCO, their instance masks are generated at a low resolution (32x32 pixels) and in practice are not often used in real-time applications.

Another branch of instance segmentation methods rely on dense-prediction, segmentation networks to generate instance masks at input resolution. Most of these methods~\cite{fathi2017semantic,newell2017associative,kong2018recurrent,de2017semantic,novotny2018semi} are based on an embedding loss function, which forces the feature vectors of pixels belonging to the same object to be similar to each other and sufficiently dissimilar from feature vectors of pixels belonging to other objects. 
Recently, works~\cite{novotny2018semi,liu2018intriguing} have shown that the spatial-invariant nature of Fully Convolutional Networks is not ideal for embedding methods and propose to either incorporate coordinate maps~\cite{liu2018intriguing} or use so-called semi-convolutions~\cite{novotny2018semi} to alleviate this problem. Nevertheless, at the current time these methods still fail to achieve the same performance as the proposal-based ones.

In light of this, a more promising and simple method is proposed by Kendall et al.~\cite{kendall2017multi}, inspired by~\cite{liang2015proposal}, in which they propose to assign pixels to objects by pointing to its object's center. This way, they avoid the aforementioned problem of spatial-invariance by learning position-relative offset vectors. Our method is based on the same concept, but integrates the post-processing clustering step directly into the loss function and optimizes the intersection-over-union of each object's mask directly. Related to our method is the very recent work of Novotny et al.~\cite{novotny2018semi}. Although similar in concepts, they use a different loss function and still apply a detection-first principle.

Also inspired by~\cite{kendall2017multi} is Box2Pix, a work proposed by Uhrig et al.~\cite{uhrig2018box2pix}, where they first predict bounding boxes based on a single-shot detection method, and then associate pixels by pointing to object centers, which can afterwards be efficiently clustered. Its focus lays on real-time instance segmentation and shows promising results on the Cityscapes dataset. Our method also shows real-time performance on the Cityscapes dataset, but at a much higher accuracy.

Our loss relaxation by learning an optimal clustering margin shows some similarites with \cite{novotny2017learning, kendall2017uncertainties}, where they integrate the aleatoric uncertainty into the loss function. In contrast to these works, we directly use the learned margin at test time.

\section{Method}
\begin{figure*}[h!]
    \begin{center}
    	\includegraphics[width=1\linewidth]{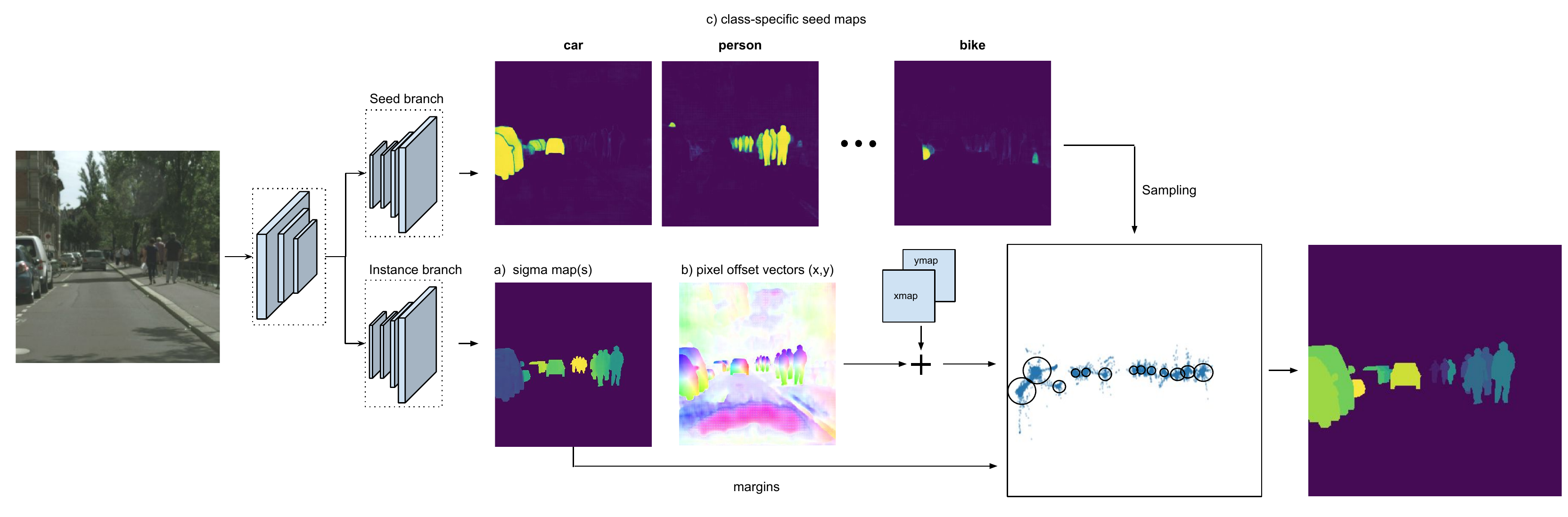}
    \end{center}
    \vspace*{-5mm}
    \caption{Instance segmentation pipeline. The bottom branch of the network predicts: a) a sigma value for each pixel, which directly translates into a clustering margin for each object. Bigger objects are more blueish, meaning a bigger margin, and smaller objects are more yellowish, meaning a smaller margin. b) Offset vectors for each pixel, pointing at the center of attraction, and displayed using a color-encoding where the color indicates the angle of the vector. The top branch predicts a seed map for each semantic class. A high value indicates that the offset vector of that pixel points directly at the object center. Notice therefore that the borders have a low value, since they have more difficulty of knowing to which center to point. The pixel embeddings (= offset vectors + coordinate vectors) and margins calculated from the predicted sigma are also displayed. The cluster centers are derived from the seed maps.}
    \label{fig:pipeline}
\end{figure*}

We treat instance segmentation as a pixel assignment problem, where we want to associate pixels with the correct objects. To this end we learn an offset vector for each pixel, pointing to its object's center. Unlike the standard regression approach, which we explain further in \ref{sub:regression}, we also learn an optimal clustering region for each object and by doing so we relax the loss for pixels far away from the center. This is explained in \ref{sub:learnable_margin}. To locate the object's centers, we learn a seed map for each semantic class, as described in \ref{sub:seed_map}. The pipeline is graphically depicted in figure~\ref{fig:pipeline}.

\subsection{Regression to the instance centroid}
\label{sub:regression}

The goal of instance segmentation is to cluster a set of pixels $ \mathcal{X} = \{ x_0,x_1,x_2,..., x_N \} $, with $x$ a 2-dimensional coordinate vector, into a set of instances $ \mathcal{S} =\{ S_0, S_1, ..., S_K \}$.

An often used method is to assign pixels to their corresponding instance centroid $C_k = \frac{1}{N} \sum_{x \in S_k} x $ . This is achieved by learning an offset vector $o_i$ for each pixel $x_i$, so that the resulting (spatial) embedding $e_i = x_i + o_i$ points to its corresponding instance centroid. Typically, the offset vectors are learned using a regression loss function with direct supervision:

\begin{equation}
    \mathcal{L}_{regr} = \sum_{i=1}^{n} \| o_i - \hat{o}_i \|
\end{equation}

where $\hat{o}_i = C_k - x_i$ for $x_i \in S_k$. However, the above method poses two issues at inference time. First, the locations of the instance centroids have to be determined and second, the pixels have to be assigned to a specific instance centroid. To solve these problems, previous methods rely on density-based clustering algorithms to first locate a set of centroids $ \mathcal{C} = \{ C_0,C_1, ..., C_K \} $ and next assign pixels to a specific instance based on a minimum distance-to-centroid metric :

\begin{equation}
    e_i \in S_k : k = \text{arg} \min_{ \mathcal{C} } \| e_i - C \|
\end{equation}

Since this post-processing step (center localization and clustering) is not integrated within the loss function, the network cannot be optimized end-to-end for instance segmentation, leading to inferior results. 

\subsection{Learnable margin}
\label{sub:learnable_margin}

The assignment of pixels to instance centroids can be incorporated into the loss function by replacing the standard regression loss with a hinge loss variant, forcing pixels to lay within a specified margin $\delta$ (the hinge margin) around the instance centroid:

\begin{equation}
    \mathcal{L}_{hinge} = \sum_{k=1}^{K} \sum_{ e_i \in S_k } \max(  \| e_i - C_k \| - \delta, 0)
\end{equation}

This way, at test time, pixels are assigned to a centroid by clustering around the centroid with this fixed margin:

\begin{equation}
    e_i \in S_k \iff \| e_i - C_k \| < \delta
\end{equation}

However, a downside to this method is that the margin $\delta$ has to be selected based on the smallest object, ensuring that if two small objects are next to each other, they can still be clustered into two different instances. If a dataset contains both small and big objects, this constraint negatively influences the accuracy of big objects, since pixels far away from the centroid will not be able to point into this small region around the centroid. Although using a hinge loss incorporates the clustering into the loss function, given the said downside it is not usable in practice.

To solve this issue we propose to learn an instance specific margin. For small instances a small margin should be used, while for bigger objects, a bigger margin would be preferred. This way, we relax the loss for pixels further away from the instance centroid, as they are no longer forced to point exactly at the instance centroid.

In order to do so, we propose to use a gaussian function $\phi_k$ for each instance $S_k$, which converts the distance between a (spatial) pixel embedding $e_i = x_i + o_i$ and the instance centroid $C_k$ into a probability of belonging to that instance:

\begin{equation}
\phi_k(e_i) = \exp\left({ - \frac{ \| e_i - C_k \| ^ 2}{ 2 \sigma_k ^ 2}  }\right)
\end{equation}

A high probability means that the pixel embedding $e_i$ is close to the instance centroid and is likely to belong to that instance, while a low probability means that the pixel is more likely to belong to the background (or another instance). More specifically, if $\phi_k(e_i) > 0.5$, than that pixel, at location $x_i$, will be assigned to instance k.

Thus, by modifying the sigma parameter of the mapping function, the margin can be controlled:

\begin{equation}
    \text{margin} = \sqrt{-2\sigma_k^2\ln0.5}
\end{equation}

A large sigma will result in a bigger margin, while a small sigma will result in a smaller margin. This additionally requires the network to output a $\sigma_i$ at each pixel location. We define $\sigma_k$ as the average of all $\sigma_i$ belonging to instance k:

\begin{equation}
    \sigma_k = \frac{1}{|S_k|} \sum_{\sigma_i \in S_k} \sigma_i
\end{equation}

Since for each instance k the gaussian outputs a foreground/background probability map, this can be optimized by using a binary classification loss with the binary foreground/background map of each instance as ground-truth. As opposed to using the standard cross-entropy loss function, we opt for using the Lovasz-hinge loss~\cite{berman2018lovasz, yu2015learning} instead. Since this loss function is a (piecewise linear) convex surrogate to the Jaccard loss, it directly optimizes the intersection-over-union of each instance. Therefore we do not need to account for the class imbalance between foreground and background.


Note that there is no direct supervision on the sigma and offset vector outputs of the network (as was the case in the standard regression loss). Instead, they are jointly optimized to maximize the intersection-over-union of each instance mask, receiving gradients by backpropagation through the Lovasz-hinge loss function and through the gaussian function.

\subsection{Intuition}

Let us first consider the case where the sigma (margin) of the Gaussian function is kept fixed. In contrast with the standard regression loss explained above, we don't have an explicit loss term pulling instance pixels to the instance centroid. Instead, by minimizing the binary loss, instance pixels are now indirectly forced to lay within the region around the instance centroid and background pixels are forced to point outside this region.

When the sigma is not fixed but a learnable parameter, the network can now also modify sigma to minimize the loss more efficiently. Aside from pulling instance pixels within the (normally small) region around the instance centroid and pushing background pixels outside this region, it can now also modify sigma such that the size of the region is more appropriate for that specific instance. Intuitively this would mean that for a big object it would adapt sigma to make the region around the centroid bigger, so that more instance pixels can point inside this region, and for small objects to choose a smaller region, so that it is easier for background pixels to point outside the region. 

\subsection{Loss extensions}
\paragraph{Elliptical margin}
In the above formulation of the gaussian function we have used a scalar value for sigma. This will result in a circular margin. However, we can modify the mapping function to use a 2-dimensional sigma:

\begin{equation}
    \phi_k(e_i) = \exp\left({ -  \frac{ ( e_{ix} - C_{kx} ) ^ 2}{ 2 \sigma_{kx} ^ 2} - \frac{ ( e_{iy} - C_{ky} ) ^ 2}{ 2 \sigma_{ky} ^ 2} }\right)
\end{equation}

By doing so, the network has the possibility of also learning an elliptical margin, which may be better suited for elongated objects such as pedestrians or trains. Note that in this case the network has to output two sigma maps, one for $\sigma_x$ and one for $\sigma_y$.

\paragraph{Learnable Center of Attraction} Another modification can be made on the center of the gaussian function. Currently, we place the gaussian in the centroid $C_k$ of each instance. By doing so, pixel embeddings are pulled towards the instance centroid. However, we can also let the network learn a more optimal \textit{Center of Attraction}. This can be done by defining the center as the mean over the embeddings of instance k. This way, the network can influence the location of the  \textit{center of attraction} by changing the location of the embeddings:

\begin{equation}
\phi_k(e_i) = \exp\left({ - \frac{ \| e_i - \frac{1}{|S_k|}\sum_{e_j \in S_k} e_j \| ^ 2}{ 2 \sigma_k ^ 2}  }\right)
\end{equation}

We will test these modifications in the ablation experiment section.

\subsection{Seed map}
\label{sub:seed_map}

At inference time we need to cluster around the center of each object. Since the above loss function forces pixel embeddings to lay close to the object's center, we can sample a \textit{good} pixel embedding and use that location as instance center. Therefore, for each pixel embedding we learn how far it is removed from the instance center. Pixel embeddings who lay very close to their instance center will get a high score in the seed map, pixel embeddings which are far away from the instance center will get a low score in the seed map. This way, at inference time, we can select a pixel embedding with a high seed score, indicating that that embedding will be very close to an object's center. 

In fact, the seediness score of a pixel embedding should equal the output of the gaussian function, since it converts the distance between an embedding and the instance center into a closeness score. The closer the embedding is laying to the center, the closer the output will be to 1. 

Therefore, we train the seed map with a regression loss function. Background pixels are regressed to zero and foreground pixels are regressed to the output of the gaussian. We train a seed map for each semantic class, with the following loss function:

\begin{equation}
    \mathcal{L}_{\text{seed}} = \frac{1}{\text{N}} \sum_i^\text{N} \mathbbm{1}_{\{s_i \in S_k\}} \| s_i - \phi_k(e_i) \|^2 + \mathbbm{1}_{\{s_i \in \text{bg}\}} \| s_i - 0 \|^2
\end{equation}

with $s_i$ the network's seed output of pixel i. Note that this time we consider $\phi_k(e_i)$ to be a scalar: gradients are only calculated for $s_i$.

\subsection{Post-processing}
At inference time, we follow a sequential clustering approach for each class-specific seed map. The pixels in the seed map with the highest value indicate which embeddings lay closest to an object's center. The procedure is to sample the embedding with the highest seed value and use that location as instance center $\hat{C_k}$. At the same location, we also take the sigma value, $\hat{\sigma}_k$. By using this center and accompanying sigma, we cluster the pixel embeddings into instance $S_k$:

\begin{equation}
    e_i \in S_k \iff \exp\left({ - \frac{ \| e_i - \hat{C_k} \| ^ 2}{ 2 \hat{\sigma_k} ^ 2}  }\right) > 0.5
\end{equation}

We next mask out all clustered pixels in the seed map and continue sampling until all seeds are masked. We repeat this process for all classes.

To ensure that during sampling $\hat{\sigma}_k \approx \sigma_k = \frac{1}{|S_k|} \sum_{\sigma_i \in S_k} \sigma_i$, we add a smoothness term for each instance to the total loss:
\begin{equation}
    \mathcal{L}_{\text{smooth}} = \frac{1}{|S_k|} \sum_{\sigma_i \in S_k} \| \sigma_i - \sigma_k \|^2
\end{equation}

\section{Experiments}
\begin{table*}[ht!]
    \small
    \centering
    \begin{tabular}[t]{l|l|c c | c c c c c c c c}
         method & training data & $AP$ & $AP_{50}$ & person & rider & car & truck & bus & train & mcycle & bicycle \\
         \hline
         DIN~\cite{arnab2017pixelwise}          & \texttt{fine\,+\,coarse}  & 23.4 & 45.2 & 20.9 & 18.4 & 31.7 & 22.8 & 31.1 & 31.0 & 19.6 & 11.7 \\ 
         SGN~\cite{liu2017sgn}                  & \texttt{fine\,+\,coarse}  & 25.0 & 44.9 & 21.8 & 20.1 & 39.4 & 24.8 & 33.2 & 30.8 & 17.7 & 12.4 \\
         PolygonRNN++~\cite{acuna2018efficient} & \texttt{fine}             & 25.5 & 45.5 & 29.4 & 21.8 & 48.3 & 21.1 & 32.3 & 23.7 & 13.6 & 13.6 \\ 
         Mask R-CNN~\cite{he2017mask}           & \texttt{fine}             & 26.2 & 49.9 & 30.5 & 23.7 & 46.9 & 22.8 & 32.2 & 18.6 & 19.1 & 16.0 \\
         GMIS~\cite{liu2018affinity}            & \texttt{fine\,+\,coarse}  & 27.6 & 44.6 & 29.3 & 24.1 & 42.7 & 25.4 & 37.2 & 32.9 & 17.6 & 11.9 \\
         PANet~\cite{liu2018path}               & \texttt{fine}             & 31.8 & 57.1 & 36.8 & 30.4 & 54.8 & 27.0 & 36.3 & 25.5 & 22.6 & 20.8 \\
         Mask R-CNN~\cite{he2017mask}           & \texttt{fine\,+\,COCO}    & 31.9 & 58.1 & 34.8 & 27.0 & 49.1 & 30.1 & 40.9 & 30.9 & 24.1 & 18.7 \\   
         PANet~\cite{liu2018path}               & \texttt{fine\,+\,COCO}    & 36.4 & 63.1 & 41.5 & 33.6 & 58.2 & 31.8 & 45.3 & 28.7 & 28.2 & 24.1 \\
         \hline
         ours           & \texttt{fine}             & 27.6 & 50.9 & 34.5 & 26.1 & 52.4 & 21.7 & 31.2 & 16.4 & 20.1 & 18.9 \\
    \end{tabular}
    \vspace{3mm}
    \caption{Results on Cityscapes $test$ set. With a score of 27.6 AP we reach second place on the benchmark, compared with the \texttt{fine}-only methods.}
    \label{tab:results_cityscapes}
\end{table*}

In this section we evaluate the performance of our instance segmentation method on the Cityscapes dataset. To find the best settings of our loss function, we first analyze the different aspects in an ablation study. Afterwards we report results of our best model on the test set of Cityscapes and compare with other top performing methods. Since our method is optimized for fast instance segmentation, we also report a time comparison with other instance segmentation methods.

\subsection{Implementation details}

\paragraph{Network architecture} We use the ERFNet-architecture~\cite{romera2018erfnet} as base-network. ERFNet is a dense-prediction encoder-decoder network optimized for real-time semantic segmentation. We convert the model into a 2-branch network, by sharing the encoder part and having 2 separate decoders. The first branch predicts the sigma and offset values, with 3 or 4 output channels depending on sigma ($\sigma$ vs $\sigma_{xy}$). The other branch outputs N seed maps, one for each semantic class. The offset values are limited between [-1,1] with a tanh activation function, sigma is made strictly positive by using an exponential activation function, effectively letting the network predict $log(\frac{1}{2\sigma^2})$.

\paragraph{Coordinate map} Since the Cityscapes images are of size 2048x1024, we construct a pixel coordinate map so that the x-coordinates are within the range of [0,2] and the y-coordinates within the range of [0,1]. This way, the difference in coordinate between two neighboring pixel is 1/1024, both in x and y direction. Because the offset vectors can have a value between [-1,1], each pixel can point at most 1024 pixels away from its current location. 

\paragraph{Training procedure} We first pre-train our models on 500x500 crops, taken out of the original 2048x1024 train images and centered around an object, for 200 epochs with a batch-size of 12. This way, we don't spend to much computation time on background patches without any instances. Afterwards we finetune the network for another 50 epochs on 1024x1024 crops with a batch-size of 2 to increase the performance on the bigger objects who couldn't fit completely within the 500x500 crop. During this stage, we keep the batch normalization statistics fixed. We use the Adam optimizer and polynomial learning rate decay $(1 - \frac{\text{epoch}}{\text{max epoch}})^{0.9}$. During pre-training we use an initial learning rate of 5e-4, which we lower to 5e-5 for finetuning. Training takes roughly 24 hours on two NVIDIA 1080 Ti GPU's. Next to random cropping, we also apply random horizontal mirroring as data-augmentation.

\begin{figure}
    \begin{center}
    	\includegraphics[width=1\linewidth]{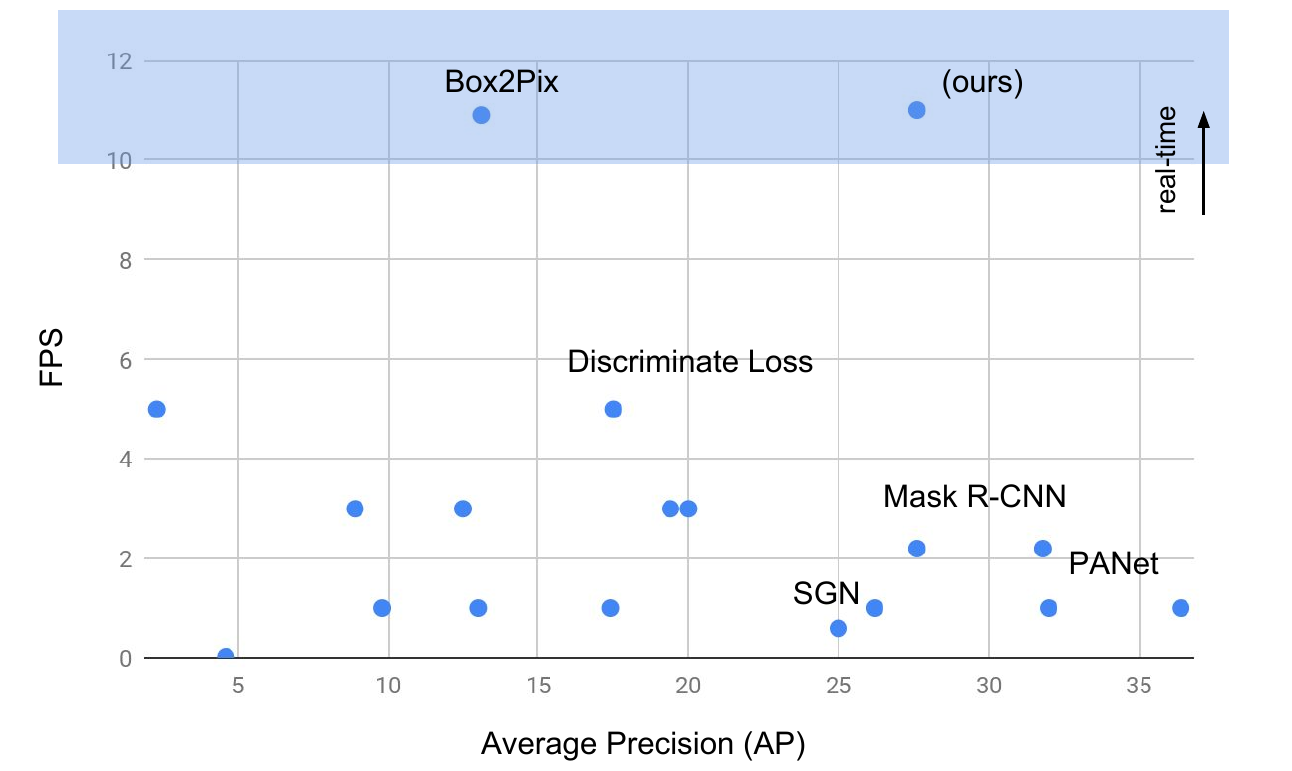}
    \end{center}
    \caption{Speed accuracy trade-off between instance segmentation methods on the Cityscapes benchmark. Our method is the first real-time method with high accuracies. Image adapted from ~\cite{uhrig2018box2pix}}
    \label{fig:timing_results}
\end{figure}

\subsection{Cityscapes dataset}
The Cityscapes dataset is high quality dataset for urban scene understanding. It consists out of 5,000 finely annotated images (\texttt{fine}) of 2048 by 1024 pixels, with both semantic and instance-wise annotations, and 20,000 coarsely annotated images (\texttt{coarse}) with only semantic annotations. The wide range in object size and the varying scene layout makes this a challenging dataset for instance segmentation methods. 

The instance segmentation task consists in detecting objects of 8 different semantic classes and generating a binary mask for each of them. The performance is evaluated by the average precision (AP) criterion on the region level and averaged over the different classes. Aside from AP, $\text{AP}_{50\%}$ for an overlap of 50\,\%, $\text{AP}_{100m}$ and $\text{AP}_{50m}$ for objects restricted to respectively 100m and 50m are also reported. 

In the following experiments we will only use the \texttt{fine} train set to train our models, which consists out of the following classes with their respective number of objects:

\vspace{3mm}

\begin{centering}
    \footnotesize
    \centering
    \begin{tabular}{c|c|c|c|c|c|c|c}
         person & rider & car & truck & bus & train & mcycle & bicycle \\
         \hline 
         17.9k & 1.8k & 26.9k & 0.5k & 0.4k & 0.2k & 0.7k & 3.7k \\
    \end{tabular}
\end{centering}

Note that some classes (truck, bus, train) are highly underrepresented, which will negatively effect the model's test performance on those specific classes.

\begin{table*}
    \small
    \centering
    \begin{tabular}{c|c|c|c c c c c c c c}
        $\sigma / \sigma_{xy}$  & CoA & $AP[\texttt{val}]_{gt}$ & person & rider & car & truck & bus & train & mcycle & bicycle \\
        \hline
        $\sigma_{fixed}$         & centroid    & 28.0 & 32.3 & 28.1 & 45.1 & 30.2 & 37.3 & 14.4 & 19.9 & 16.9 \\
        $\sigma$                  & centroid    & 38.7 & 36.4 & 33.6 & 54.5 & 42.7 & 56.0 & 36.7 & 24.9 & 24.5 \\
        $\sigma$                  & learnable   & 39.5 & 39.4 & 35.4 & 56.0 & 40.3 & 57.6 & 34.6 & 26.1 & 26.5 \\
        $\sigma_{xy}$           & centroid    & 39.1 & 38.0 & 33.9 & 54.5 & 42.0 & 59.4 & 37.8 & 23.0 & 24.5 \\
        $\sigma_{xy}$           & learnable   & 40.5 & 39.3 & 34.5 & 55.5 & 44.3 & 59.8 & 41.2 & 24.8 & 25.0 \\
    \end{tabular}
    \vspace{3mm}
    \caption{Ablation experiments evaluated on the Cityscapes validation set using a ground-truth sampling approach. We measure the performance of a fixed sigma, the difference in using a scalar vs. 2-dimensional sigma and the difference in using the centroid or learnable center as \textit{center of attraction}.}
    \label{tab:abl_exp_gt}
\end{table*}

\subsection{Ablation Experiments}

In this section we evaluate the influence of the different parameters of our loss function on the validation set of Cityscapes: we investigate the importance of a learnable sigma, the difference in using the instance centroid or a learnable center as the \textit{center of attraction}, and the difference in using a scalar or a 2 dimensional sigma. Since we want to measure the effect on the instance part, we remove the object detection and classification part from the equation by using the ground truth annotations to localize the objects and assigning the correct semantic class, which is indicated in the tables as $\text{AP}_{gt}$.

\paragraph{Fixed vs. learnable sigma}
In this experiment we evaluate the importance of a learnable, instance-unique sigma over a fixed one. As explained in section~\ref{sub:learnable_margin}, when using a fixed sigma, the value has to be selected based on the size of the smallest object we still want to be able to separate, and is therefore set to correspond with a margin of 20 pixels. The results can be seen in table~\ref{tab:abl_exp_gt}. The significant performance difference (28 AP vs. 38.7 AP) shows the importance of having a unique, learnable sigma for each instance. Notice also that for classes with relatively more small instances, the difference is less pronounced, as expected.

\paragraph{Fixed vs. learnable Center of Attraction}
As described in the method section, the \textit{center of attraction} (CoA) of an instance can be defined as either the centroid, or more general, as a learnable center calculated by taking the mean over all spatial embeddings belonging to the instance. Intuitively, by giving the network the opportunity to decide on the location of the CoA itself, it can learn a more optimal location than the standard centroid. In table~\ref{tab:abl_exp_gt} we evaluate the two different approaches on the Cityscapes validation set using a ground-truth sampling approach,both in the case of a scalar or a 2-dimensional sigma. As predicted, in both cases we achieve a higher AP-score when using a learnable center instead of the fixed centroid, with a noticeable improvement over all classes. 

\begin{figure}
    \begin{center}
    	\includegraphics[width=1\linewidth]{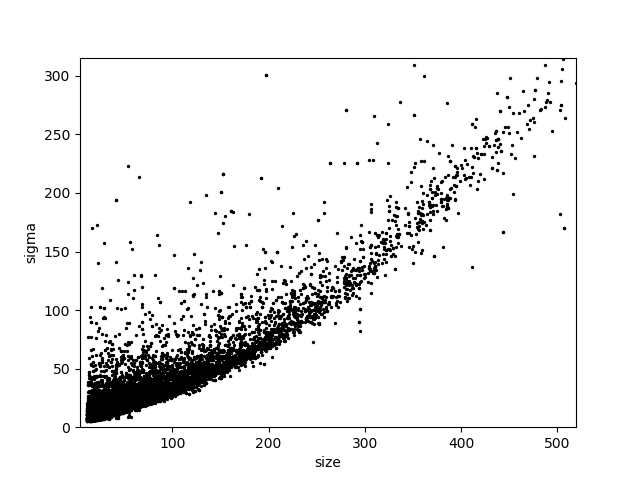}
    \end{center}
    \caption{Learned margin against the object's size. Each dot represents an object in the dataset. As predicted, we notice a positive correlation between the margin and the object's size.}
    \label{fig:sigma_vs_size}
\end{figure}

\paragraph{Circular vs. elliptical margin}
The margin for each instance is defined by the learnable sigma parameter in the gaussian function. This sigma can either be a scalar ($\sigma$), which results in a circular margin, or a two-dimensional vector ($\sigma_{xy}$), resulting in an elliptical margin. For rectangular objects (e.g.~pedestrians) a circular margin is not optimal, since it can only expand until it reaches the shortest border. An elliptical margin however would have the possibility to stretch and adapt to the shape of an object, possibly resulting in a higher accuracy. In table~\ref{tab:abl_exp_gt} we compare both methods and verify that a 2-dimensional sigma (elliptical margin) indeed performs better than a scalar one (circular margin).

Since sigma is a learnable parameter, we have no direct control over its value. Intuitively, since sigma controls the clustering margin, we speculated that for big objects sigma will be bigger, resulting in a bigger margin, and smaller for small objects. To verify this, in figure~\ref{fig:sigma_vs_size} we plotted sigma in function of the object's size. As predicted, there is indeed a positive correlation between an object's size and sigma.

\begin{figure*}
    \begin{center}
    	\includegraphics[width=1\linewidth]{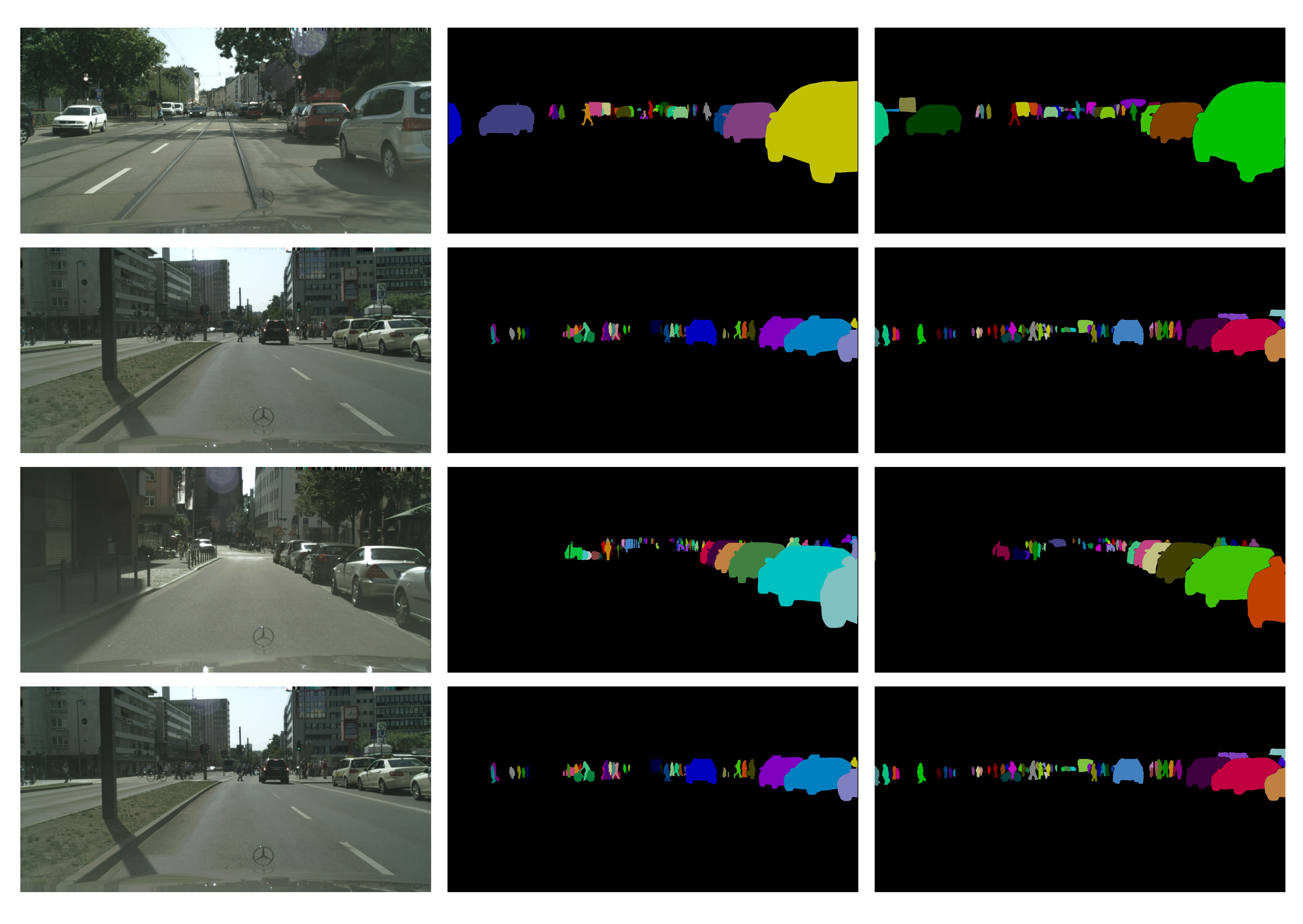}
    \end{center}
    \caption{Results on the Cityscapes dataset. From left to right: input image, ground-truth and our predictions. Notice that our method is very good at detecting small objects and often predicts more correct objects than annotated in the ground-truth.}
    \label{fig:results}
\end{figure*}

\subsection{Results on Cityscapes}
In table~\ref{tab:results_cityscapes} we report results on the Cityscapes test set and compare with other high performing methods. Note however that it is important to pay attention at the training data on which a method is trained. Since the truck, bus and train classes are highly underrepresented in the \texttt{fine} set, methods who only train on this set will perform less on these classes than methods who augment their dataset with the \texttt{coarse} or \texttt{COCO} set. 

Comparing our method against the other \texttt{fine}-only methods, we occupy the second place with an AP-score of 27.6, locating ourselves between between the popular Mask R-CNN (26.2) and PANet(31.8). Notice however that we do much better on the person (34.5 vs 30.5), rider (26.1 vs 23.7) and car class (52.4 vs 46.9) than Mask R-CNN. If we compare our method with GMIS, a method trained on both the \texttt{fine} and \texttt{coarse} set, we notice that although it has the same AP-score as our method, it only performs better on the truck, bus and train class (because of the extra \texttt{coarse} set) and performs worse on all other classes. 

Although it is not fair to compare our method against methods trained on \texttt{fine+COCO}, we do notice that we achieve similar results on person (34.5 vs 34.8) and rider (26.1 vs 27.0), and even perform better on car (52.4 vs 49.1) and bicycle (18.9 vs 18.7) with respect to Mask R-CNN.

\begin{table}
    \small
    \centering
    \begin{tabular}{l|c|c|c}
        Method & AP & $\text{AP}_{50}$ & FPS \\
        \hline 
        Deep Contours~\cite{van2016instance}     & 2.3  & 3.6  & 5 \\
        Box2Pix~\cite{uhrig2018box2pix}          & 13.1 & 27.2 & 10.9 \\
        BAIS~\cite{hayder2017boundary}           & 17.4 & 36.7 & $<$1 \\
        Discriminate Loss~\cite{de2017semantic}  & 17.5 & 35.9 & 5 \\
        DWT~\cite{bai2017deep}                   & 19.4 & 35.3 & $<3$ \\
        Dynamic Net~\cite{arnab2017pixelwise}    & 20.0 & 38.3 & $<3$ \\
        SGN~\cite{liu2017sgn}                    & 25.0 & 44.9 & 0.6 \\
        Mask-RCNN~\cite{he2017mask} (fine)       & 26.2 & 49.9 & 2.2 \\
        PANet~\cite{liu2018path}                 & 31.8 & 57.1 & $<$1 \\
        \hline
        ours                                     & 27.6 & 50.9 & \textbf{11} \\
    \end{tabular}
    \vspace{3mm}
    \caption{Approximate timing results of instance segmentation methods on a resolution of 2048x1024 with test set accuracy ~\cite{uhrig2018box2pix}. Methods which are either to slow or have a very low accuracy are left out.}
    \label{tab:timing_results}
\end{table}

\subsection{Timing}
In table~\ref{tab:timing_results} we compare the execution speed of different methods. This is also depicted in fig~\ref{fig:timing_results}. Up to this moment, most methods have put there focus on accuracy rather than on execution speed. Mask-RCNN (26.2 AP - 1fps) and derivatives have high accuracy, but slow execution speed. Other methods,like Discriminative loss (17.5 AP - 5fps) or Box2Pix (13.1 AP - 10.9fps) achieve higher frame rates by downsampling resolution or using single shot detection methods, but dramatically lack behind in accuracy compared to Mask R-CNN. Since our method is based on the ERFNet network and combined with a clustering loss function, we are the first ones to achieve high accuracy  combined with real time performance (27.6 AP - 11fps). More specifically, the forward pass at a resolution of 2MP takes 65ms and the clustering step requires 26ms.

\section{Conclusions}
In this work we have proposed a new clustering loss function for instance segmentation. By using a gaussian function to convert pixel embeddings into a foreground/background probability, we can optimize the intersection-over-union of each object's mask directly and learn an optimal, object-specific clustering margin. We show that when applied to a real-time, dense-prediction network, we achieve top results on the Cityscapes benchmark at more than 10 fps, making our method the first proposal-free, real-time instance segmentation method with high accuracy.

{\bf Acknowledgement:} This work was supported by Toyota, and was carried out at the TRACE Lab at KU Leuven (Toyota Research on Automated Cars in Europe - Leuven).


{\small
\bibliographystyle{ieee}
\bibliography{egbib}
}

\end{document}